\documentclass[english]{ccdconf}
\usepackage{color}
\usepackage[linkcolor=red,citecolor=blue]{hyperref}
\usepackage[justification=centering]{caption}
\usepackage{array}
\usepackage{booktabs}
\usepackage{threeparttable}
\usepackage{enumerate}
\usepackage{amsthm,amssymb}
\begin{document}

\title{An ensemble learning framework based on group decision making}
\author{Jingyi He\aref{csu}, Xiaojun Zhou\aref{csu,peng},
      Rundong Zhang\aref{csu}, Chunhua Yang\aref{csu}}

\affiliation[csu]{School of Automation,
Central South University, Changsha, 410083}

\affiliation[peng]{the Peng Cheng Laboratory,
Shenzhen, 518000
        \email{michael.x.zhou@csu.edu.cn}}

\maketitle

\begin{abstract}
Classification problem is a significant topic in machine learning which aims to teach machines how to group together data by particular criteria. In this paper, a framework for ensemble learning (EL) method  based on group decision making (GDM) has been proposed to resolve this issue. In this framework, base learners can be considered as decision makers, different categories can be seen as alternatives, classification results obtained by diverse base learners can be considered as performance ratings, and the precision, recall and accuracy which can reflect the performances of the classification methods can be employed to identify the weights of decision makers in GDM. Moreover, considering that the  precision and recall defined in binary classification problem can not be used directly in  multi-classification problem, the One vs Rest (OvR) has been proposed to obtain the precision and recall of the base learner for each category. The experimental results demonstrate that the proposed EL method based
on GDM has higher accuracy than other 6 current popular classification methods in most instances, which verifies the effectiveness of the
proposed method.
\end{abstract}


\footnotetext{This work was supported by the National Natural
Science Foundation of China (Grant No. 61860206014, 61873285), the 111 Project(Grant No. B17048), the Innovation-Driven Plan in Central South University (Grant No. 2018CX012), Hunan Provincial Natural Science Foundation of China (Grant No. 2018JJ3683).}

\section{Introduction}
Classification is a significant field of data mining. A classifier is constructed by a tagged training set. It is used to judge the class label for the testing set which is unlabeled. There are many machine learning methods that can be used to solve classification problems. However, there is no such a machine learning method which can perform better than others in classification overall. To solve this problem, ensemble learning (EL) method could be adopted, which is commonly regarded as the machine learning interpretation for the wisdom of the crowd \cite{1999Liu}. Learners which compose an ensemble are usually called base learners, and their results are generally combined by majority voting. The main premise of EL method is that by combining multiple models, the errors of a base learner will likely be compensated by base learners, and as a result, the overall prediction performance of the ensemble would be better than that of a base learner \cite{2015Yi}. Many studies have demonstrated that EL method could be used to solve classification problems. A new classification method based on EL method for real-time detection of seizure activity in epilepsy patients was discussed by Hosseini  et al. \cite{2018Hosseini}. Mesquita et al. \cite{2017Mesquita} proposed ensemble methods for classification and regression using minimal learning machine. In recent years, class imbalance becomes a pressing issue,
and some researches have shown that EL method can clearly enhance the classification performance in this respect \cite{2018Bi,2019Chakraborty}.

It is common for us to make a choice that taking various suggestions into consideration.
The process of making a choice with taking many suggestions from our friends or
experts into account is group decision making (GDM) \cite{1999Li}. GDM is a common
occurrence of every aspect in our daily life. For example, shopping
 online considering our friends' experience. GDM is a process of finding an optimal alternative that has the highest degree of satisfaction from a set of alternatives considering various
advisors' suggestions. And current literatures mainly address
the combination of GDM with other discipline domains \cite{2015Ali,2005Ye}. Ure$\tilde{n}$a et al. \cite{2019Kou} analyzed the trust and reputation in distributed networked scenarios in decision making to reach consensus. Dong et al. \cite{2019Dong} proposed a novel framework that hybridizes both the process of making closer opinions realized by consensus reaching processes and the evolving relationships among experts based on social network analysis.

As mentioned above, EL method is a comprehensive term for methods that combine multiple inducers to make a decision, typically in supervised machine learning tasks. The combining strategy in EL method is especially important, but EL method combining with GDM has not been studied fully yet. In this paper, a framework based on EL method and GDM has been proposed to settle classification problems. Experiments are carried out with  5 public data sets and 6 different base learners to compare their performance. Extensive experiments have verified that the EL method combining with GDM is  better than most of other machine learning algorithms. It also  demonstrates that the framework combining EL method based on GDM is effective.\\
The remainder of this paper is organized as follows. In Section \ref{section2}, a new EL method called EL based on GDM will be proposed. In Section \ref{section3}, the experimental settings and the performance of the proposed method comparing with other methods will be presented. Finally, the conclusion and future work will be drawn in Section \ref{section4}.
\section{Ensemble learning based on group decision making}\label{section2}
In this section, the framework for EL method based on GDM will be presented. First, how classical EL methods work and
their differences will be introduced. Second, the GDM model will be illustrated. Finally, the framework for EL method based on GDM will be given in detail.
\subsection{Classical ensemble learning  methods}\label{section2.1}
Ensemble methods can be divided into two main frameworks: one is the dependent; the other is the independent \cite{2018O}.
 In the former, the output of each base learner works on the model of the next base learner. In the latter, each base learner works independently from other base learners \cite{2010Liu}. A brief review of two popular ensemble methods of both frameworks has been performed in sections below.

\textbf{Random Forest(RF)} is a typical representative of the independent framework. It consists of many independent decision trees, each tree is generated by the different training sets. In RF, a training set is produced by sampling with replacement \cite{2016M}. In order to produce a random tree which is still sufficiently accurate, RF has one difference from the classic decision tree. Before splitting each decision tree, the best feature should be selected from a feature set which is generated by random selection. The decision trees’ combination strategy for RF is majority vote.

\textbf{AdaBoost} is another widely used independent framework. The main idea of AdaBoost is that the next base learner is focused on the instances which are incorrectly classified by last base learner. This procedure is implemented by modifying the weights of instances \cite{2019Jiang}. The initial weights of all instances are the same. After each iteration, the weights of the correct instances will be reduced. Instead, the weights of the error instances will be increased. Besides, weights are also allocated to the base learners based on their overall predictive performance. Same as RF, weighted voting is adopted. In this case, each base learner has a different emphasis, and combining all of the base learners can allow the model to focus on all instances.\\
In general, 5 differences in generation of base learners could be summarized as follows:
\begin{itemize}
  \item[$\blacksquare$] Sample selection: Random Forest takes sampling with replacement, but AdaBoost uses the same training set for each iteration.
  \item[$\blacksquare$] Weights: Random Forest has the same sample weights and base learner weights, but AdaBoost has the different sample weights and base learner weights for each iteration.
  \item[$\blacksquare$] Parallel computing: Each tree in a Random Forest can be computed in parallel, but base learners in AdaBoost can only serial generated.
  \item[$\blacksquare$] Error: Random Forest focuses on reducing variance, but AdaBoost tends to reducing bias.
  \item[$\blacksquare$] Diversity: The diversities of base learners are ensured by different training sets in Random Forest, but AdaBoost is done by changing the weights of the instances.
\end{itemize}

It is obvious that the classical EL methods change the samples to produce different models. In this paper, 6 different machine learning methods could be used to achieve this goal.

\subsection{Group decision making methods}
Considering the GDM problem of ranking $m$ alternatives, represented as $A_{1},A_{2},\ldots,A_{m}$ based on the descending order from `best' to `worst'  \cite{1999Li}. Forming a committee of $K$ decision makers, denoted as $E_{1},E_{2},\ldots,E_{K}$ to identify $n$ decision criteria, called $C_{1},C_{2},\ldots,C_{n}$. Each decision maker can evaluate every alternatives and its corresponding criteria individually, then the performance ratings of every alternatives and its corresponding criteria, called $x_{ij}^{k}(i=1,2,\ldots,n;j=1,2,\ldots,m;k=1,2,\ldots,K)$ will be obtained. Also, the important weight of each criterion and each decision maker can be denoted as $w_{1},w_{2},\ldots,w_{n}$ and $W_{1},W_{2},\ldots,W_{K}$ with respect to some overall objectives, respectively \cite{2019Kou,2019Dong,2019Liu}. Performance ratings and importance weights assigned by the decision makers could be crisp numbers ranging from $0$ to $1$.\\
The performance rating $x_{ij}^{k}$ assigned to alternative $A_{i}$ by decision maker $E_{k}$ for criterion $C_{j}$ can be used to measure how well $A_{i}$ satisfies $C_{j}$ for decision maker $E_{k}$ \cite{2000Chen}. Then the decision making data can be collected to form the matrices $X_{k}$:

\begin{equation}
X_{k}=\left[
  \begin{array}{ccccc}
    x_{11}^{k} & \cdots & x_{1j}^{k} & \cdots & x_{1m}^{k}\\
    \vdots & \ddots & \vdots & \ddots & \vdots\\
    x_{i1}^{k} & \cdots & x_{ij}^{k} & \cdots & x_{im}^{k}\\
    \vdots & \ddots & \vdots & \ddots & \vdots\\
    x_{n1}^{k} & \cdots & x_{nj}^{k} & \cdots & x_{nm}^{k}\\
  \end{array}
\right]
\label{matrices}
\end{equation}
The decision making method can be used to aggregate these matrices with importance weights to make the decision making process more conveniently \cite{2013Wan}.
In this work, GDM is used for information fusion of the base learners in EL method.
\subsection{A framework for ensemble learning based group decision making}
In fact, multiple different classification methods can obtain absolutely different classification results even for solving the same classification problem. It is hard to find a overall best classification method. In order to get the best classification result obtained by  combining multiple classification methods under the classification problem, a framework for EL method based on GDM can be employed. In EL method, each classification method called base learner. In this framework, base learners can be considered as decision makers, different categories can be seen as alternatives, and different classification results obtained by diverse base learners will be described as performance ratings.
So, for a multi-classification problem which has $m$ categories, the decision making data can be collected to form the matrices $X_{k}$. The matrices $X_{k}$ can be written as Eq. (\ref{matrices}).
When $K$ base learners are used in EL method and they evaluate each category in n dimensions, the importance weights of each criterion can be represented as $w=[w_{1},w_{2},\ldots,w_{n}]$ and the importance of the weight of each decision maker can be represented as $W=[W_{1},W_{2},\ldots,W_{K}]$.
The precision, recall and accuracy can reflect the performance of the classification method. In this study, a combination of these indexes is used, rather than a single index, to measure the performance of classification methods. The precision, recall and accuracy are respectively given from Eq. (\ref{precision}) to  Eq. (\ref{accuracy}).
\begin{figure*}
  \centering
  \includegraphics[width=12cm]{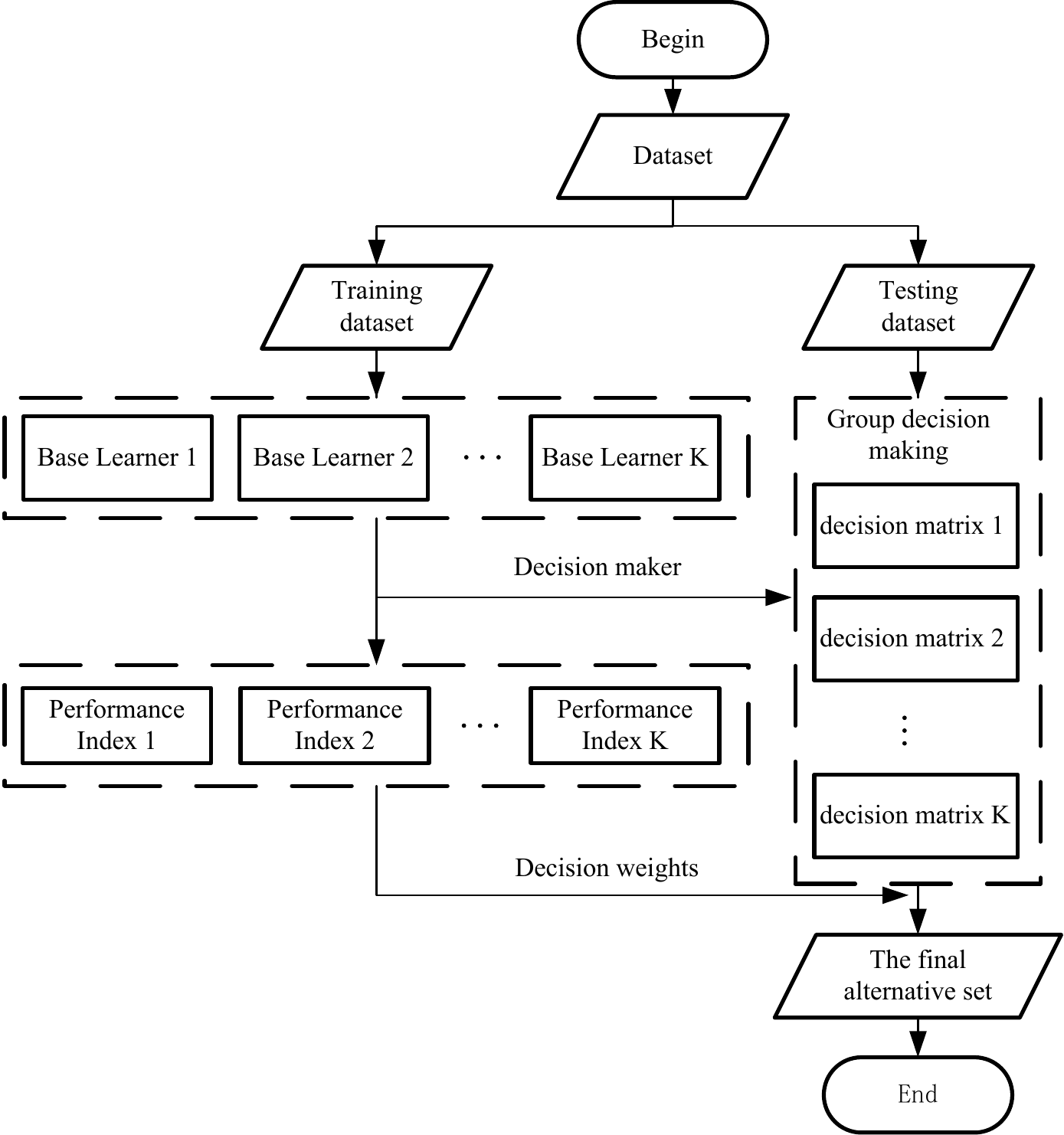}\\
  \caption{The framework for ensemble learning method based on group decision making.}\label{fig_framework}
\end{figure*}
\begin{table}[!htbp]
\caption{Classification result confusion matrix.}
\centering
\setlength{\tabcolsep}{3pt}
\begin{tabular}{p{2.5cm}<{\centering} p{2cm}<{\centering} p{1cm}<{\centering} p{2cm}<{\centering} }
  \toprule
  Truth & \quad & Prediction & \quad  \\
  \cmidrule(ll){2-4}
  \quad &Positive Class&\quad & Negative Class\\
  \midrule
  Positive Class & $TP$&\quad & $FN$ \\
  Negative Class & $FP$&\quad  & $TN$ \\
  \bottomrule
\end{tabular}
\label{tab_dataset}
\end{table}

\begin{equation}
P=\frac {TP}{TP+FP}
\label{precision}
\end{equation}
\begin{equation}
R=\frac {TP}{TP+FN}
\label{recall}
\end{equation}
\begin{equation}
A=\frac {TP+TN}{TP+FP+TN+FN}
\label{accuracy}
\end{equation}
where $P,R,A$ are the precision, recall and accuracy. A class could be identified as positive class and the rest of classes belong to negative class.\\
The precision is the number of correct positive results divided by the number of positive results predicted by the classifier. The recall is the number of correct positive results divided by the number of all relevant samples (all samples that should have been identified as positive). The accuracy is the ratio of number of correct predictions to the total number of input samples. The precision and recall are a pair of contradictory.
 For example, if all the instances are predicated to be positive class, the recall must be equal to one, but the precision will be very low. If there are K classification methods, the precision, recall and accuracy can be described as $P=[P_{1},P_{2},...,P_{K}],R=[R_{1},R_{2},...,R_{K}],A=[A_{1},A_{2},...,A_{K}]$, respectively. The P, R, A of each base learner can be identified as the components of P, R, A. If the dataset has a total of $m$ categories, the components of $P,R,A$ are defined as $P_{k}={[p_{1},p_{2},...,p_{m}]}^{T}$, $R_{k}={[r_{1},r_{2},...,r_{m}]}^{T}$, $A_{k}={[a_{1},a_{2},...,a_{m}]}^{T}(k=1,2,...,K)$, where $p_{i},r_{i},a_{i}$ represents the precision, recall and accuracy of base learner $k$ for category $i$. Based on the above description, the weight of each decision maker is defined as follows:
 \begin{equation}
W_{k}=P_{k}+R_{k}+A_{k}\quad(k=1,2,...,K)
\label{Wk}
\end{equation}
 From the Eq. (\ref{Wk} ), it is obviously that with larger indicators come larger weights.
Finally, classification results should be calculated as:
 \begin{equation}
H(x)=\arg\max\sum_{k=1}^{K}X_{k}W_{k}(x)
\end{equation}
where $H(x)$ can be regarded as the alternative $x$ with the highest score.\\
The framework for EL method based on GDM is illustrated in Fig. \ref{fig_framework}. As can be seen from this framework, a dataset is divided into training set and testing set. A training set is used to train models and the performance indexes can be obtained. Each model is a base learner. GDM has been introduced to fuse the information of these base learners. Decision makers give each alternative evaluating scores which are called performance ratings. The decision matrixes can be formed according to these performance ratings. The weights of  decision makers are related to its performance indexes. In this framework, GDM is used to determine which alternative is the most suitable one.
\section{Experimental Results}\label{section3}
In this section, several examples are illustrated to show the
effectiveness of our proposed method.
\begin{table*}[!htbp]
\caption{The description of classification data sets.}
\centering
\setlength{\tabcolsep}{3pt}
\begin{tabular}{p{1.8cm}<{\centering} p{1.8cm}<{\centering} p{1.8cm}<{\centering} p{1.8cm}<{\centering} p{1.8cm}<{\centering} p{1.8cm}<{\centering} p{4.5cm}<{\centering}}
  \toprule
  DataSets & Instances & Classes & Attributes & Train & Test & Size of classes \\
  \midrule
  CMC & 1473 & 3 & 9 & 1179 & 294 & 629 ,\quad333,\quad511 \\[2pt]
  Glass & 214 & 6 & 9 & 172 & 42 & 29 ,\quad76,\quad70,\quad17,\quad13,\quad19 \\[2pt]
  Seeds & 210 & 3 & 7 & 168 & 42 & 70 ,\quad70,\quad70 \\[2pt]
  Sonar & 208 & 2 & 60 & 167 & 41 & 97 ,\quad111 \\[2pt]
  Wine & 178 & 3 & 13 & 143 & 35 & 59 ,\quad71,\quad48 \\
  \bottomrule
\end{tabular}
\label{tab_dataset}
\end{table*}
In this paper, experiments are carried out with 5 real-world data sets which are all available at the UCI Machine Learning Repository (\url{http://archive.ics.uci.edu}). Details about these datasets can be found in Table \ref{tab_dataset}.
The table lists the number of instances, classes, features and instances of per class for each dataset.
Specific tasks for the datasets are listed as follows:
\begin{enumerate}[(i)]
  \item The Contraceptive Method Choice Dataset(CMC): The problem is to predict the current contraceptive method choice of a woman based on her demographic and socio-economic characteristics.
  \item The Glass Identification Dataset(Glass): The purpose of this is determining whether the glass is a type of "float" glass or not.
  \item The Seeds Dataset(Seeds): The instances belonging to three different varieties of wheat, the goal is figuring out what kind of wheat each instance is.
  \item The Connectionist Bench Dataset(Sonar): The dataset contains signals obtained from a variety of different aspect angles. Whether the sample is a mine or a rock is judged by this information.
  \item The Wine Dataset(Wine): These data are the results of a chemical analysis of wines. Which kind of wine the instances belong to should be predicted.
\end{enumerate}
First, 80$\%$ of each data set have been randomly selected as the training set and the rest as the testing set. By this way, each dataset is divided into two subsets. In this paper, 6 common classification methods are employed to settle the same classification problem. That is to say, we have 6 base learners which are Support Vector Machine(SVM) \cite{2018Yu}, Back-Propagation(BP) Neural Network \cite{2019Han}, Logistic Regression(LR) \cite{2019Hoang}, k-Nearest Neighbor(KNN) \cite{2016Aburomman}, Random Forest(RF), Extreme Learning Machine(ELM) \cite{2014Samat}. These classification methods have been widely used to solve so many kinds of different classification problems. These classification methods have been employed to ensure the diversity of base learners in this paper. They will be briefly introduced as follows:
\begin{enumerate}[(1)]
  \item SVM: It plots each instance as a point in a $n$-dimensional space (where $n$ is the number of features), and the value of each feature is the value of a specific coordinate. SVM is trying to find a hyperplane that divides the dataset into two categories.  It needs some treatments for multi-classification problems.
  \item BP: A neural network is an algorithm that endeavors to recognize underlying relationships in a set of data through a process that mimics the way the human brain operates. BP Neural Network is an algorithm widely used in the training of feedforward neural networks for supervised learning.
  \item LR: The core of the regression method is to find the most appropriate parameters for the function $f(x)$, so that the value of the function and the sample are closed. LR fits a function whose value reflects the probability of sample belonging to its class in probability.
  \item KNN: It is a simple algorithm that stores all available cases and categorizes new cases of its $K $ neighbors by majority votes. For instance, a good way to understand a stranger maybe get information from his friends.
  \item RF: For more details, see the Section \ref{section2.1}.
  \item ELM: ELM is regarded as an improvement on feedforward neural network and back propagation algorithm. The characteristic of ELM is that the weight of hidden layer nodes is randomly or artificially given. Also, the weight does not need to be updated and the learning process only calculates the output weight.
\end{enumerate}

For each base learner, each training set can be used to train a model which can be used to classify the testing set. The model's ability to categorize testing sets determines its weight. Taking CMC dataset as a motivating example, a training set is used to train 6 base learners, called BP, ELM, LR, SVM, FR, KNN. These base learners can be seen as 6 decision makers and 3 categories can be represented as 3 alternatives. According to the GDM, each decision maker assigns a score for every alternative. So 6 decision matrices can be shown as follows:
\begin{equation}
    X_{k}=[x_{11}^{k} , x_{12}^{k} , x_{13}^{k}]\\
\end{equation}
where $X_{k}$ is the decision matrix of decision maker $k$.\\
In order to obtain the performance indexes of the decision maker, the precision, recall and accuracy are calculated. In view of three categories, the components of $P,R,A$ are $P_{k}={[p_{1},p_{2},p_{3}]}^{T},R_{k}={[r_{1},r_{2},r_{3}]}^{T},A_{k}={[a_{1},a_{2},a_{3}]}^{T}$,
respectively. Their concrete subvalues are given in the Table \ref{tab_example}.
\begin{table}[!htbp]
\caption{Performance indicators of base learners($\%$).}
\centering
\setlength{\tabcolsep}{3pt}

\begin{tabular}{p{1.5cm}<{\centering} p{1.5cm}<{\centering} p{1.4cm}<{\centering} p{1.4cm}<{\centering} p{1.4cm}<{\centering}}
  \toprule
   Base  & \quad & \quad & Classes & \quad \\
    \cmidrule(ll){3-5}
   learners & Indicators & First & Second & Third \\
  \midrule
  \quad & Precision & 84.40 & 50.00 & 71.19  \\
  SVM & Recall & 92.00 & 68.00 & 84.00 \\
  \quad & Accuracy & 51.19 & 51.19 & 51.19 \\
  \quad & Precision & 89.19 & 65.71 & 81.54  \\
  BP & Recall & 90.83 & 69.70 & 84.13 \\
  \quad & Accuracy &59.30 & 90.83 & 59.30 \\
  \quad & Precision & 83.33 & 44.44 & 65.52  \\
  LR & Recall & 93.45 & 69.56 & 84.44 \\
  \quad & Accuracy & 52.20 & 52.20 & 52.20 \\
  \quad & Precision & 75.73 & 46.81 & 60.94  \\
  KNN & Recall & 88.64 & 68.75 & 79.59 \\
  \quad & Accuracy & 47.12 & 47.12 & 47.12 \\
  \quad & Precision & 86.13 & 52.50 & 73.24  \\
  RF & Recall & 95.93 & 80.77 & 91.23 \\
  \quad & Accuracy & 64.75 & 64.75 & 64.75 \\
  \quad & Precision & 80.19 & 57.14 & 36.36  \\
  ELM & Recall & 64.39 & 37.33 & 20.34 \\
  \quad & Accuracy & 42.37 & 42.37 & 42.37 \\
  \bottomrule
\end{tabular}
\label{tab_example}
\end{table}

\begin{figure}
  \includegraphics[width=8.5cm]{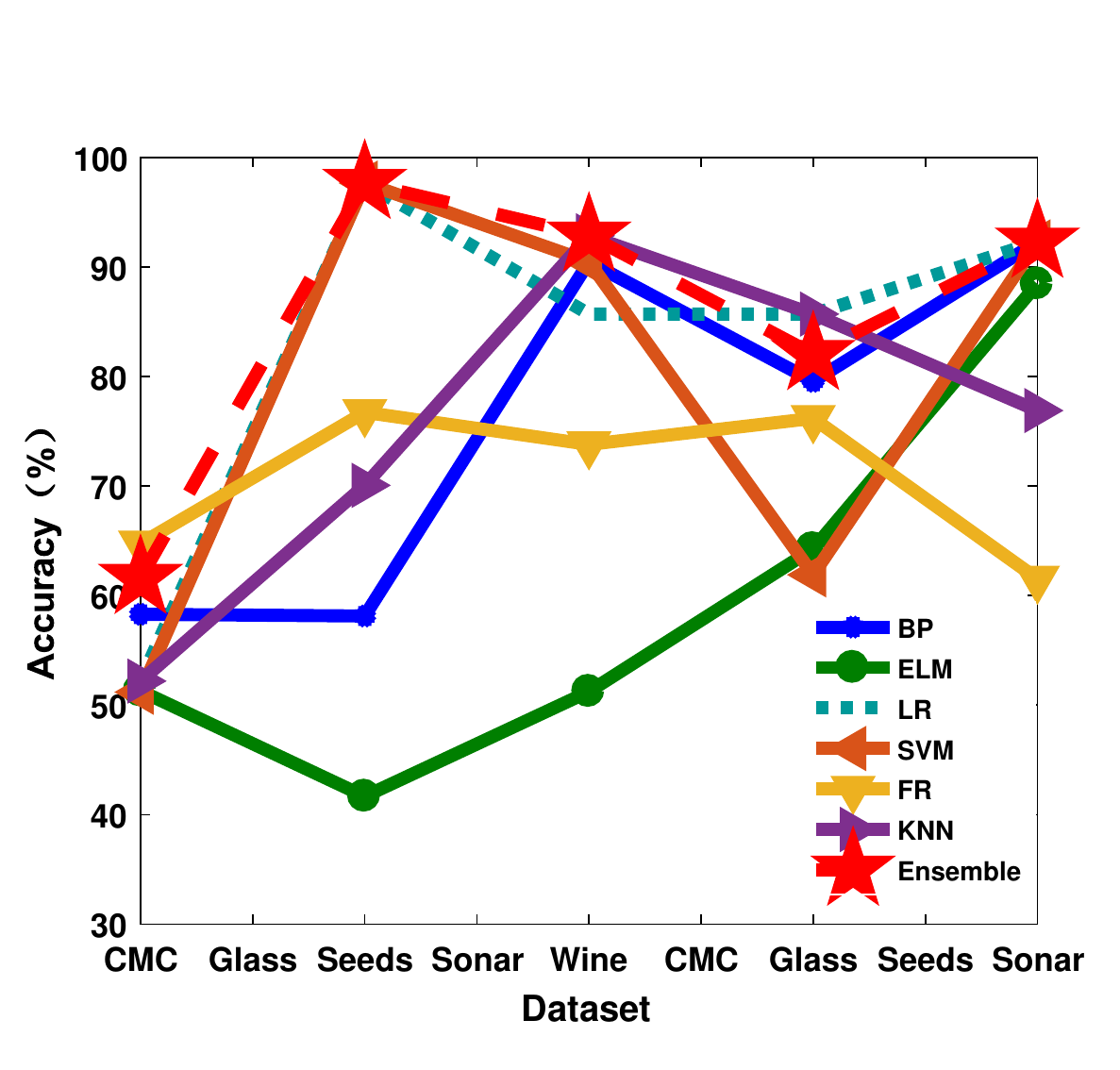}\\
  \caption{Comparison of experimental results.}\label{fig_result}
\end{figure}

At last, classification results are calculated as Eq. (\ref{W},\ref{H}).
\begin{equation}
W_{k}=P_{k}+R_{k}+A_{k}\quad(k=1,2,...,6)
\label{W}
\end{equation}
\begin{equation}
H(x)=\arg\max\sum_{k=1}^{6}X_{k}W_{k}(x)
\label{H}
\end{equation}
where $W_{k}$ is the weight of decision maker $k$.\\
Subsequent results are listed in Table \ref{Accuracy} and Fig. \ref{fig_result}. The performance of the base learners are compared with that of  EL method based on GDM proposed in this paper. In Table \ref{Accuracy}, the highest accuracy has been noted for each dataset. As one can notice, considering the recently proposed EL method based on GDM, our method achieved higher classification rates in most of the data sets, and it is never the worst of all the base learners. It is very easy to find that our proposed method outperforms other algorithms in solving this classification problems from Fig. \ref{fig_result}.

\begin{table*}[!htbp]
\linespread{1.5}
\caption{Performance comparison with base learners for classification($\%$).}
\centering
\setlength{\tabcolsep}{3pt}
\begin{tabular}{p{1.925cm}<{\centering} p{1.925cm}<{\centering} p{1.925cm}<{\centering} p{1.925cm}<{\centering} p{1.925cm}<{\centering} p{1.925cm}<{\centering} p{1.925cm}<{\centering} p{1.925cm}<{\centering}}
  \toprule
  \quad & BP & ELM & LR & SVM & FR & KNN & Ensemble \\
  \midrule
  CMC & 58.3 & 49.49 & 52.2 & 51.19 & \textbf{64.75} & 52.2 & 61.6 \\[2pt]
  Glass & 58.14 & 34.88 & \textbf{97.67} & \textbf{97.67} & 76.74 & 79.07 & \textbf{97.67} \\[2pt]
  Seeds & 90.48 & 45.24 & 85.71 & 90.48 & 73.81 & \textbf{92.86} & \textbf{92.86} \\[2pt]
  Sonar & 79.57 & 64.27 & \textbf{85.71} & 61.90 & 76.19 & \textbf{85.71} & 82.50 \\[2pt]
  Wine & \textbf{92.3} & 88.46 & \textbf{92.3} & \textbf{92.3} & 61.54 & 76.90 & \textbf{92.3} \\
  \bottomrule
\end{tabular}
\begin{tablenotes}
\item[1] The numbers in boldface indicate the highest accuracy among all methods
\end{tablenotes}
\label{Accuracy}
\end{table*}

\section{Conclusion and Future Work}\label{section4}
Multiple different classification methods can obtain absolutely different classification results even for solving the same classification problem. A framework for EL method based on GDM has been proposed to obtain the consistent result. In EL method, each classification method is called base learner. In this paper, base learners can be considered as decision makers, different categories can be seen as alternatives, different classification results obtained by diverse base learners will be described as performance ratings, and the precision, recall and accuracy which can reflect the performance of the classification method can be employed to identify the weights of decision makers in GDM. 6 current popular machine learning methods have been employed to settle the same classification problem. 5 classical classification datasets have been adopted to conduct the experiment. The experimental results demonstrate that our proposed EL method based on GDM has the highest accuracy than other 6 current popular classification methods, which verifies that our proposed method is effective. In general, the strategy and method this paper proposed are efficient in dealing with the classification problem.  In the future, there are three more perspectives can be further studied. The first is that more criteria of alternatives can be considered in the framework. The second is to utilize improved GDM method in this framework to settle the classification problem. The third is adopting the EL method based on GDM to solve real-world problems in industrial process.
\balance

\end{document}